\documentclass[11pt]{article}

\usepackage[T1]{fontenc}
\usepackage[utf8]{inputenc}

\usepackage[preprint]{acl}

\usepackage{times}
\usepackage{latexsym}
\usepackage{booktabs}
\usepackage{multirow}
\usepackage{subcaption}
\usepackage{tabularx}
\usepackage{array}
\usepackage{xcolor}

\usepackage[T1]{fontenc}

\usepackage[utf8]{inputenc}

\usepackage{microtype}

\usepackage{inconsolata}
\usepackage[most]{tcolorbox}
\usepackage{graphicx}
\usepackage{hyperref}
\usepackage{url}
\usepackage{amsmath}
\usepackage{amssymb}
\usepackage{CJKutf8}
\title{Language Lives in Sparse Dimensions: Toward Interpretable and Efficient Multilingual Control for Large Language Models}

\author{
Chengzhi Zhong\thanks{This denotes equal contribution.}\textsuperscript{1},\hspace{1em}
Fei Cheng\footnotemark[\value{footnote}]\textsuperscript{1}, \hspace{1em}
Qianying Liu\textsuperscript{2}, \hspace{1em} \\
\textbf{Yugo Murawaki\textsuperscript{1},} \hspace{1em}
\textbf{Chenhui Chu\textsuperscript{1},} \hspace{1em} 
\textbf{Sadao Kurohashi\textsuperscript{1,2} }\\
\textbf{\textsuperscript{1}} Kyoto University, Japan \hspace{1em}
\quad
\textbf{\textsuperscript{2}} NII, Japan \hspace{1em}
\\
\texttt{zhong@nlp.ist.i.kyoto-u.ac.jp} \\
\texttt{\{feicheng, murawaki, chu, kuro\}@i.kyoto-u.ac.jp} \\
\texttt{ying@nii.ac.jp} \\
}

\begin{document}
\maketitle

\begin{abstract}

Large language models exhibit strong multilingual capabilities despite limited exposure to non-English data. 
Prior studies show that English-centric large language models map multilingual content into English-aligned representations at intermediate layers and then project them back into target-language token spaces in the final layer. From this observation, we hypothesize that this cross-lingual transition is governed by a small and sparse set of dimensions, which occur at consistent indices across the intermediate to final layers. Building on this insight, we introduce a simple, training-free method to identify and manipulate these dimensions, requiring only as few as 50 sentences of either parallel or monolingual data. Experiments on a multilingual generation control task reveal the interpretability of these dimensions, demonstrating that the interventions in these dimensions can switch the output language while preserving semantic content, and that it surpasses the performance of prior neuron-based approaches at a substantially lower cost. Our code is publicly available.\footnotemark[1]

\end{abstract}
\footnotetext[1]{\url{https://github.com/ku-nlp/language-specific-dimensions}}

\begin{figure}[htb]
  \centering
  \includegraphics[width=1.0\linewidth]{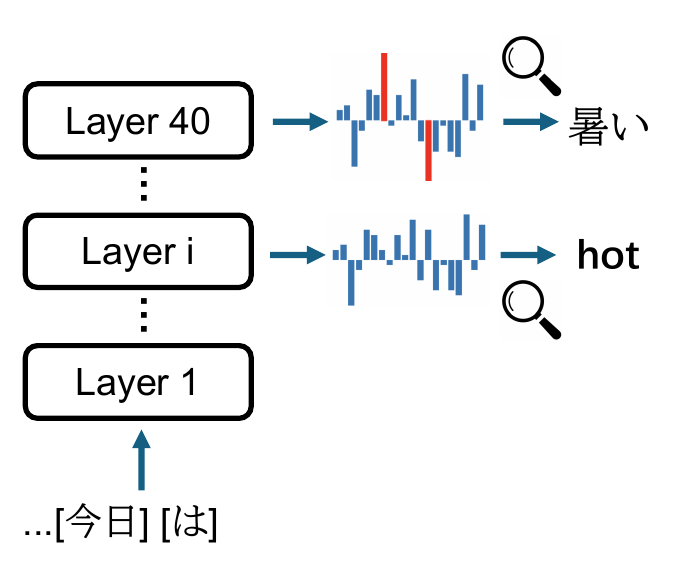}
  \caption{Hypothesis of language-specific dimensions.
  The logit lens 
  shows that English-centric LLMs often surface the English answer in intermediate layers before producing the target-language answer.
  The prompt is \begin{CJK}{UTF8}{goth}``English:"Today is hot."-日本語:"今日は''\end{CJK}.
  We ask the model to predict the omitted word ``hot'' in Japanese. Dimensions in red are likely to be language-specific.}
  \label{fig:intro}
\end{figure}

\section{Introduction}

Large language models (LLMs), though predominantly trained on English-centric corpora, have nevertheless demonstrated strong multilingual capabilities even with limited exposure to non-English data, enabling a wide range of multilingual applications.  
Recent studies~\cite{latent}  have shown that
English-centric LLMs map multilingual content into English-aligned representations at intermediate layers, before projecting them back into target-language token spaces in the later layers.

Various studies have sought to uncover the internal mechanisms of this transfer by investigating how specific components of LLMs manage language-dependent behavior such as vocabulary and grammar.
Neuron-level studies~\cite{neuron3,neuron_kojima,neuron_tang} analyze the activation patterns of feed-forward network units across large-scale multilingual corpora, identifying language-specific neurons whose manipulation can steer generation toward a target language. Sparse-Autoencoder (SAE) based approaches~\cite{saes} instead train massive sparse-autoencoders on large corpora to isolate high-dimensional features tied to language choice. 


While these approaches examine which LLM components encode language identity, our perspective is to take a more intrinsic view of multilingual transfer through the lens of the trajectory of representations, tracing how representations transform across layers.
As highlighted in red in Fig.~\ref{fig:intro}, we examine the representation trajectories from intermediate to final layers, and reveal that the transition follows a distinct and identifiable pattern, with sparse distributions concentrated in a small set of dimensions.
We put forward the hypothesis that the model leverages language-specific dimensions to intervene in the transition from a shared, language-agnostic representation space to language-specific token spaces. 
While most semantic content remains stable in language-agnostic dimensions, these language-specific dimensions consistently control the projection into the target token space. Moreover, these language-specific dimensions are not arbitrary; they appear in similar dimensions across different layers of the model, with later layers amplifying their magnitude. These observations suggest that cross-lingual transfer relies on adjustments along a consistent set of dimensions rather than diffuse redistribution.

In this paper, we present an in-depth analysis of language-specific dimensions, and the insights lead us to develop a method for cross-lingual transfer that manipulates these dimensions. Our method identifies these dimensions with as few as \(\sim\)50 sentences, in contrast to neuron-based methods that require logging activations over millions of tokens or SAE-based approaches that train massive auxiliary autoencoders. The approach is training-free and can operate with either limited parallel data or even only monolingual data, leveraging the model’s own cross-lingual mapping ability to perform low-resource translation. 

Concretely, we compute the representation for each language over a small corpus of as few as \(\sim\)50 sentences, and select the top-$K$ dimensions with the largest absolute differences between target language representation against English representation. 
In the inference stage, we overwrite these language-specific dimensions with the corresponding representation values, scaled by a coefficient, while leaving all other dimensions unchanged. 
By applying this single-layer intervention, our method can switch the output language to the desired target while preserving the underlying semantic content.
We evaluate the approach across multiple models, including Llama2~\cite{llama2}, Llama3.1~\cite{llama3.1}, and Aya23~\cite{aya23}, across five languages—French, German, Spanish, Chinese, and Japanese. The results confirm the existence of language-specific dimensions and show that our method outperforms neuron-based methods at steering the model to the desired target language.

In summary, our contribution is three-fold:
\begin{enumerate}
    \item We provide a systematic analysis of language-specific dimensions in LLMs, showing that they play a central role in controlling the projection from a shared, language-agnostic space into language-specific token spaces.
    \item We introduce a simple, training-free approach to identify and manipulate these dimensions using only a small amount of monolingual data (\(\sim\)50 sentences), enabling efficient and explainable control of output language without large-scale logging or auxiliary training.
    \item Through multilingual generation control experiments, we show that our method outperforms prior neuron-level approaches and could be further improved with parallel data.
\end{enumerate}

\section{Related Work}
\subsection{Multilingual LLMs}
Multilingual processing has been a goal since the earliest transformer-based language models such as multilingual BERT~\cite{mbert} and mBART~\cite{mbart}. The most direct route to multilingual capability remains large-scale pretraining on mixed-language corpora, systems such as Llama2, 3.1~\cite{llama2,llama3.1} and Aya23~\cite{aya23} can handle dozens of languages and achieve strong results on multilingual benchmarks. As various studies scale multilingual pretraining to boost performance, understanding how pretrained models carry out multilingual tasks, their internal representational factors, remains a key research direction.

\begin{figure*}[htb]
  \centering
  \includegraphics[width=\linewidth]{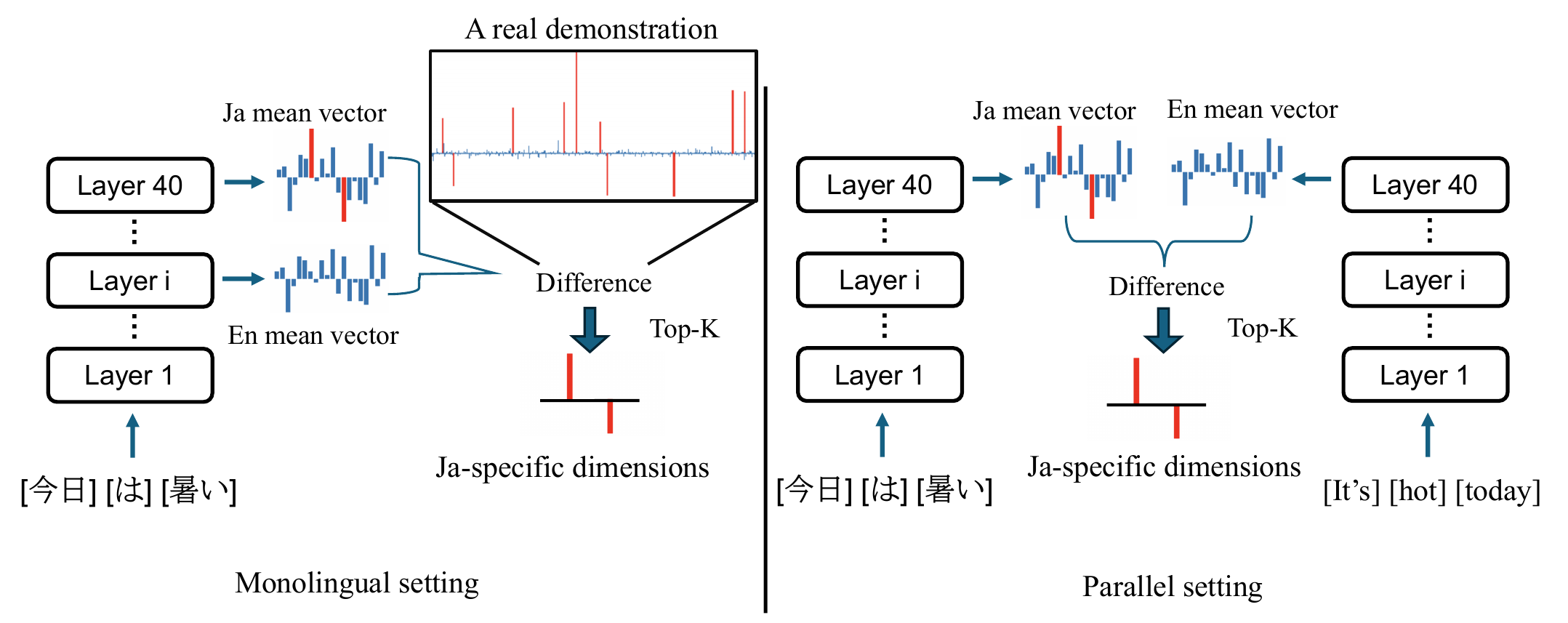}
  \caption{Two settings for identifying language-specific dimensions. Using Japanese as an example, we first average token representations within each sentence to obtain a sentence-level mean vector. Then, compare the Japanese and English means and keep the top-$K$ dimensions with the largest absolute differences as Japanese-related dimensions. We demonstrate a real difference example of LLaMA2-13B on the left top.}
  \label{fig:identify}
\end{figure*}

\subsection{Interpreting Multilingualism in LLMs}
Prior work~\cite{latent} finds a layerwise progression in LLMs such as Llama2, where early layers encode general lexical and semantic features, middle layers representations tend to reside in an English-dominant space, with later layers shifting toward target language space, depending on prompt. In English-centric models such as Llama2,  several studies~\cite{represent1,represent2} train projection metrics that map representations from English space into target language spaces, by leveraging parallel corpora to obtain representations of the same semantic content across language spaces inside the model. Applying these mappings at inference can steer the output language and improve multilingual behavior.

Other methods, including FFN-neuron manipulation~\cite{neuron_kojima,neuron_tang,neuron3} and SAE-based interventions~\cite{saes}, also enable language control. However, like the foregoing approaches, they generally rely on large datasets, frequently parallel data, which are difficult to obtain in many languages and domains, thereby limiting applicability. Thus, we seek a simple, training-free and highly data-efficient method that can achieve this objective.


\section{Methodology}\label{sec:method}

We first  propose a method to identify language-specific dimensions.
Given data availability constraints, we propose two scenarios to identify language-specific dimensions. (1) In Section~\ref{sec:mono} monolingual setting, we compare representations from an intermediate layer with those from the final layer to identify language-specific dimensions. (2) In Section~\ref{sec:para} parallel setting, we compare final layer representations for English and the target language to identify language-specific dimensions.
Then, we apply intervention to verify their effectiveness, with the setup described in Section~\ref{sec:intervene}. We modify only these dimensions at a chosen layer to steer the model toward the desired language while keeping language-agnostic semantics approximately stable. 
The procedure is described in Fig.~\ref{fig:identify}.

\subsection{Preliminary}
As a preliminary step, we investigate whether the distribution of dimensions involved in the transition from intermediate to final layers exhibits systematic patterns related to language. Specifically, we analyze the Llama2-13B model, focusing on the transition from intermediate layer 22 to the final layer 40, as shown in Fig.~\ref{fig:identify}. Following \citet{latent}, we apply logit-lens and observe that layer 22 already begins to decode semantic content in English, indicating that the core semantics are already established at this stage. When comparing the representations of the intermediate layer with those of the final layer, the majority of dimensions exhibit only minor changes, whereas a small and sparse subset, as marked red in the actual difference demonstration Fig.~\ref{fig:identify}, shows pronounced differences. 
Notably, we find that the same sparse set of dimensions also emerges when comparing later layers ($>$22) to the final layer, indicating that these spike indices are consistent across depth.
This sparse ``spike'' distribution coincides with the logit-lens observation that the decoded representation shifts from English concepts to concepts in the target language. These preliminary findings motivate our hypothesis that language-specific dimensions govern the transition from a language-agnostic space to language-specific token spaces, and they form the foundation of our method for systematically identifying and manipulating such dimensions.

\subsection{Identify Language-specific Dimensions}\label{sec:iden}

To identify language-specific dimensions, we compare target-language representations with a unified representation of the same underlying semantics, and localize the dimensions exhibiting sharp differences. In the monolingual setting, we contrast the final-layer representation with that of an intermediate layer. In the parallel setting, given that the intermediate layer representation distribution aligns closely with English, we leverage parallel sentences to obtain paired representations in English and the target language. 

\subsubsection{Monolingual setting}\label{sec:mono}

First, we treat the LLM as an encoder. For each target-language sentence, the model produces token-level representations at every layer.  We compute the mean of token-level representations at (i) a chosen intermediate layer and (ii) the final layer, yielding two sentence vectors per input. We then average these sentence vectors across all sentences to obtain corpus-level mean vectors for the intermediate and final layers, denoted by \( \boldsymbol{\mu}^{(\text{lang})}_{\ell} \in \mathbb{R}^d \) and \( \boldsymbol{\mu}^{\text{lang}}_{L} \in \mathbb{R}^d \), respectively. Finally, we compute the absolute difference per dimension
\begin{align}
\boldsymbol{\delta}^{(\text{lang)}}_{\text{diff}} \;=\; \bigl| \boldsymbol{\mu}^{(\text{lang})}_{L} - \boldsymbol{\mu}^{(\text{lang})}_{\ell} \bigr| \in \mathbb{R}^d,
\end{align}
and select the top-\(K\) indices
\begin{align}
\mathcal{I}_{K}^{(\text{lang})} \;=\; \operatorname{TopK}\!\bigl(\boldsymbol{\delta}^{(\text{lang})}_{\text{diff}},\, K\bigr).
\end{align}
These dimensions are considered language-specific and their indices are recorded for the intervention step.


\subsubsection{Parallel setting}\label{sec:para}
With parallel data, we directly compare the final layer representations of parallel sentences across English and target language. For each parallel pair, 
we collect final layer sentence representations as in the monolingual setting. We then compute the corpus-level means at the final layer $L$ for English and another for the target language, denoted by \( \boldsymbol{\mu}^{(\text{en})}_{L} \in \mathbb{R}^d \) and \( \boldsymbol{\mu}^{(\text{lang})}_{L} \in \mathbb{R}^d \), respectively.
Then, we select the top-$K$ dimensions with the largest absolute differences between the two language means use the same strategy as monolingual setting; these dimensions are taken as the language-specific dimensions for the target language, we record their index \(\mathcal{I}_{K}^{(\text{lang})} \)for the intervention step. 


\subsection{Inference-time Interventions}\label{sec:intervene}

Given the identified language-specific dimensions, we perform inference-time interventions to validate the effectiveness of these dimensions. We overwrite the values of the identified dimensions to switch the output language to the desired target while preserving semantic content. These interventions can evaluate our central hypothesis that language-specific dimensions exist in the representations of the model.

We run inference on English inputs and intervene in a chosen intermediate layer $j$. Let $\mathbf{h}_{j}\in\mathbb{R}^{d}$ denote the representation at layer $j$, and let $\mathcal{I}^{(\text{lang})}_{K}$ be the top-$K$ index set for target language $lang$, with corpus-level means $\boldsymbol{\mu}^{(lang)}_{L}\in\mathbb{R}^{d}$. Both $\mathcal{I}^{(\text{lang})}_{K}$ and $\boldsymbol{\mu}^{(lang)}_{L}$ are precomputed in Section~\ref{sec:iden}. We overwrite the selected dimensions with a scaled corpus-level mean value and keep the rest unchanged:
\begin{equation}
\mathbf{h}'_{j}[i]=
\begin{cases}
\alpha * \boldsymbol{\mu}^{(lang)}_{L}[i], & \text{if } i\in \mathcal{I}_{K}^{(\text{lang})},\\[4pt]
\mathbf{h}_{j}[i], & \text{otherwise.}
\end{cases}
\end{equation}
Here $\alpha$ is a layer-dependent scaling coefficient controlling the intervention strength (later layers typically use larger $\alpha$).



\section{Experimental Settings}
\subsection{Models}
We evaluate several models, including Llama2–7B, Llama2–13B, Llama3.1–8B, and Aya23–8B. These models come from different LLM families, are trained on different corpora, and vary in size. The Llama2 family models are predominantly English-centric, whereas the others include more multilingual data. Llama2–13B consists of 40 Transformer layers with a hidden size of 5,120. The other models use 32 layers with a hidden size of 4,096. All models are open-sourced on Hugging Face. 

\subsection{Data for Identification}
We evaluate five English-to-X directions, including English to French, German, Spanish, Chinese and Japanese.
We randomly sample 50 parallel sentence pairs from the development (dev) set of the FLORES-200~\cite{flores} for identifying language-specific dimensions. 

\subsection{Multilingual Generation Control}\label{sec:task}
We follow ~\citet{neuron_kojima} and adopt the multilingual generation control task for comparison. In this task, the model is prompted with a specially designed machine translation–like instruction:
``Translate an English sentence into a target language. English: \{source text\} Target language:.'' We then apply interventions with our method to steer the output to a desired language. Because the target language is not explicitly specified, this setting tests whether the identified language-specific dimensions can control the output language. Also, we use BLEU~\cite{bleu} to assess whether the intervention preserves the source meaning while changing the output language.

For evaluation, we follow \citet{neuron_kojima} and use the same 100 sampled test sentences per translation direction from each dataset of FLORES-200, IWSLT2017~\cite{iwslt}, and WMT~\cite{wmt14,wmt16,wmt18} and compute the mean performance. If a dataset does not contain the given direction (e.g., IWSLT2017 lacks English to Spanish), it is omitted and the average is taken over the remaining.

We evaluate with a pipeline of three metrics: 
\begin{itemize}
\setlength\itemsep{0em}
  \item Accuracy (ACC): measures the probability that the intervention successfully induces generation in the target language.
  \item BLEU: evaluates translation quality for successful samples only.
  \item ACC*BLEU: a composite score that reflects both success rate and translation quality.
\end{itemize}

For calculating accuracy, we use the fastText~\cite{fasttext} language identification classifier to detect the output language. A sample is counted as successful if the classifier’s score exceeds a threshold of 0.5~\cite{neuron_kojima}.

\subsection{MLQA}\label{sec:qa}

We add MLQA~\cite{mlqa} as an additional task. MLQA is an extractive question answering benchmark, where the model is required to answer a question by copying a short span from the provided context.
In this work, we extract 440 parallel examples from the MLQA dataset covering German, Spanish, and Chinese.

During evaluation, we use the following fixed English QA prompt format:
\begin{tcolorbox}[
    colback=gray!30,
    colframe=gray!100,
    boxrule=0.8pt,
    arc=2pt,
    left=6pt,
    right=6pt,
    top=4pt,
    bottom=4pt
]
Answer the question using a short phrase directly copied from the context.\\
Do not explain or generate additional text.\\
Context: \{context\}\\
Question: \{question\}\\
Answer:
\end{tcolorbox}

Although the prompt is always in English, we apply different intervention methods to steer the model to generate answers in a target language.
The generated answers are evaluated against gold answers in the corresponding target language.
We report performance using the F1 and Exact Match (EM) metrics.

\section{Results and Discussion}

\subsection{Pre-examination on language-specific dimension identification}

\begin{figure}[!t]
  \centering
  \includegraphics[width=0.9\linewidth]{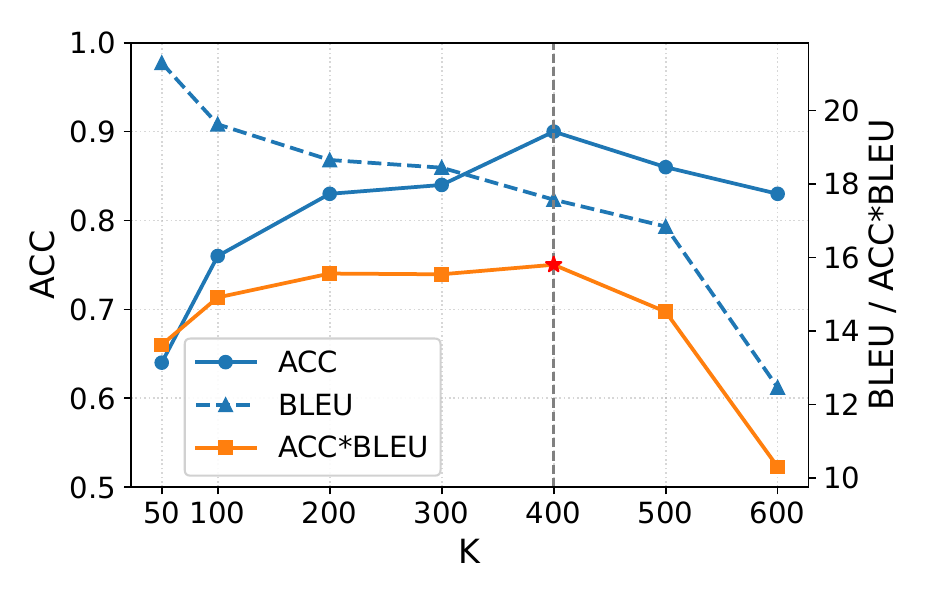}
  \caption{Selecting a top-$K$ threshold for detecting language-specific dimensions. Ablation study conducted on Llama2-7B in the parallel setting.}
  \label{fig:ablationk}
\end{figure}

We first search for an appropriate top-$K$ threshold for selecting language-specific dimensions. 
Therefore, we run the multilingual generation control experiment on Llama2-7B, applying an intervention at layer 19 with $\alpha\!=\!0.4$ in the parallel setting while varying $K$. As shown in Fig.~\ref{fig:ablationk}, as $K$ increases, ACC improves but BLEU drops. The composite ACC*BLEU improves from 13.62 at $K\!=\!50$, to 14.91 at $K\!=\!100$ (+1.29), and to 15.80 at $K\!=\!400$ (+2.18). Considering this trade-off, the most essential language-specific dimensions appear to be captured at around $K\!=\!100$, which yields a smaller loss in semantic fidelity while still improving accuracy. However, to maximize performance, we set $K\!=\!400$ for the subsequent experiments, which is about 7.8\% of the hidden size for Llama2-13B and 9.8\% for the other models.

\begin{figure}[t]
  \centering
  \includegraphics[width=0.9\linewidth]{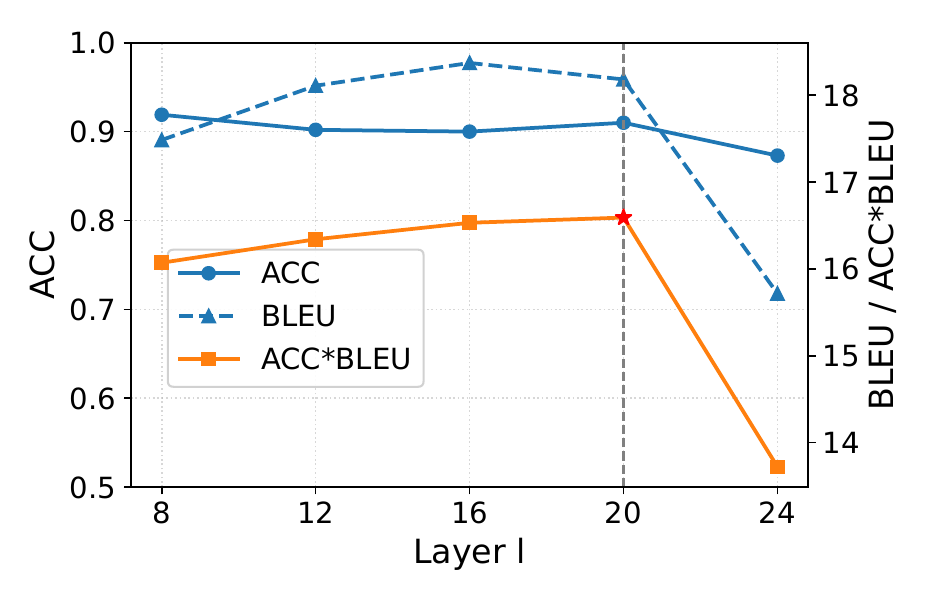}
  \caption{Choosing an anchor layer for monolingual setting. Ablation study conducted on Llama2-7B in the monolingual setting.}
  \label{fig:inter}
\end{figure}

In the monolingual setting, we need to choose an intermediate reference layer to compare with the final layer. We therefore set $K\!=\!400$, apply an intervention at layer 19 with $\alpha\!=\!0.4$, and evaluate Llama2-7B on the multilingual generation control task while varying the layer $l$ from which dimensions are identified. As shown in Fig.~\ref{fig:inter}, ACC and BLEU remain stable over a broad range of $l$, and the composite ACC*BLEU attains its best value near $l\!=\!20$ but degrades at very deep layers. This indicates that the results are not highly sensitive to the choice of intermediate layer, and we use $l\!=\!20$ in subsequent experiments.

\begin{table}[t]
\centering
\setlength{\tabcolsep}{6pt}
\renewcommand{\arraystretch}{1.15}
\begin{tabular}{lrrrrr}
\toprule
      & \textbf{zh} & \textbf{ja} & \textbf{fr} & \textbf{es} & \textbf{de} \\
\midrule
\textbf{zh} & 400 & \textbf{193} & 105  & 100  & 118  \\
\textbf{ja} &  & 400 & 88  & 98  & 113  \\
\textbf{fr} &   &   & 400 & \textbf{152} & 143 \\
\textbf{es} &   &   &  & 400 & 140 \\
\textbf{de} &   &   &  &  & 400 \\
\bottomrule
\end{tabular}
\caption{Overlap matrix of top-400 language-specific dimensions across languages for Llama2-7B in the monolingual setting. Bold numbers indicate language pairs that share more dimensions.}
\label{tab:overlap}
\end{table}

After identifying language-specific dimensions, we examine whether different languages share them. Table~\ref{tab:overlap} confirms clear cross-language overlap. Typologically closer pairs share more dimensions: Chinese and Japanese share 193 of 400 dimensions. Romance and Germanic languages also overlap more with one another. Specifically, French shares 152 and 143 dimensions with Spanish and German, respectively, substantially more than with Chinese (100) and Japanese (98). These results indicate that many language-specific dimensions are partially shared across languages rather than being unique to a single language. 

\begin{figure}[t]
  \centering
  \includegraphics[width=0.8\linewidth]{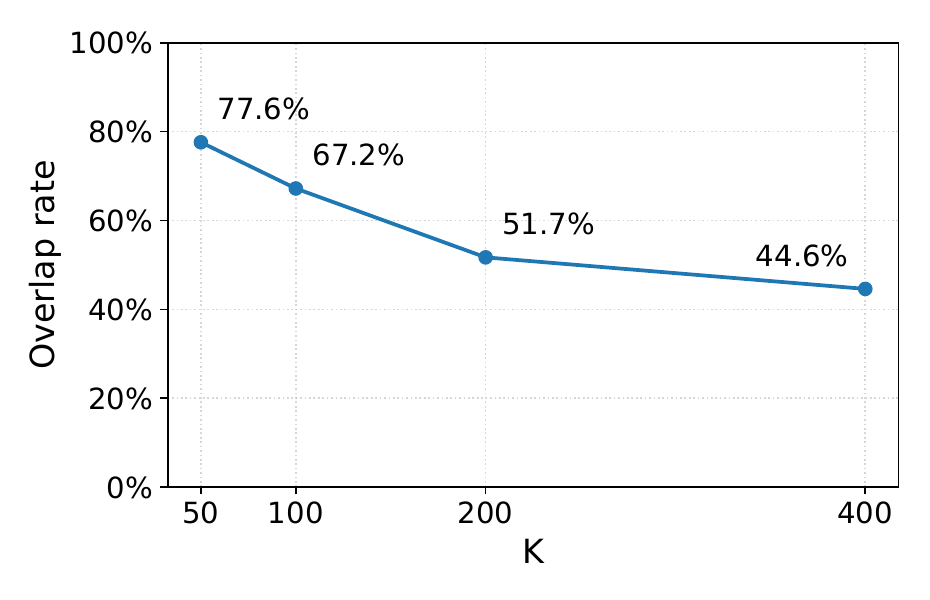}
  \caption{Overlap rate of language-specific dimensions identified by monolingual and parallel settings for different $K$. Experiments are conducted on Llama2-7B.}
  \label{fig:overlap2}
\end{figure}

\begin{table*}[t]
\centering
\setlength{\tabcolsep}{2pt}
\renewcommand{\arraystretch}{1.5}
\resizebox{\textwidth}{!}{
\begin{tabular}{ll *{2}{c} *{2}{c} *{2}{c} *{2}{c} *{2}{c} *{3}{c}}
\toprule
\textbf{Model} & \textbf{Method}
& \multicolumn{2}{c}{\textbf{Fr}}
& \multicolumn{2}{c}{\textbf{De}}
& \multicolumn{2}{c}{\textbf{Zh}}
& \multicolumn{2}{c}{\textbf{Ja}}
& \multicolumn{2}{c}{\textbf{Es}}
& \multicolumn{3}{c}{\textbf{Overall}} \\
\cmidrule(lr){3-4}\cmidrule(lr){5-6}\cmidrule(lr){7-8}\cmidrule(lr){9-10}\cmidrule(lr){11-12}\cmidrule(lr){13-15}
&
& \textbf{ACC} & \textbf{BLEU}
& \textbf{ACC} & \textbf{BLEU}
& \textbf{ACC} & \textbf{BLEU}
& \textbf{ACC} & \textbf{BLEU}
& \textbf{ACC} & \textbf{BLEU}
& \textbf{ACC} & \textbf{BLEU} & \textbf{A*B} \\
\midrule
\multirow{4}{*}{Llama2-7b}
& \textbf{\shortstack[l]{Neuron Kojima}}
& 42.0 & 15.3 & 55.7 & 15.1 & 81.3 & 10.2 & 57.5 & 7.5 & 77.0 & 16.6 & 60.8 & 12.5 & 7.56 \\
& \textbf{\shortstack[l]{Neuron Tang}}
& 83.7 & \textbf{28.2} & 72.7 & \textbf{21.1} & 86.7 & \textbf{17.9} & 20.5 & \textbf{14.4} & 98.0 & \textbf{20.8} & 72.3 & \textbf{21.8} & 15.80 \\
& \textbf{\shortstack[l]{Ours (Monolingual)}}
& 98.3 & 23.4 & 97.3 & 19.4 & 84.7 & 16.4 & \textbf{76.5} & 9.9 & 98.7 & 17.30 & 91.1 & 18.3 & \textbf{16.63} \\
& \textbf{\shortstack[l]{Ours (Parallel)}}
& \textbf{99.1} & 22.7 & \textbf{98.8} & 18.3 & \textbf{88.3} & 17.2 & \textbf{76.5} & 9.4 & \textbf{99.3} & 17.4 & \textbf{92.6} & 17.9 & 16.57 \\
\midrule
\multirow{4}{*}{Llama2-13b}
& \textbf{\shortstack[l]{Neuron Kojima}}
& 15.7 & 7.3 & 30.3 & 9.5 & 96.0 & 15.9 & 64.5 & 8.6 & 14.0 & 4.3 & 47.4 & 12.2 & 5.80 \\
& \textbf{\shortstack[l]{Neuron Tang}}
& 83.7 & \textbf{32.5} & 14.7 & \textbf{23.5} & \textbf{99.3} & \textbf{19.1} & 64.5 & 8.6 & 42.0 & \textbf{25.8} & 63.7 & \textbf{22.4} & 14.23 \\
& \textbf{\shortstack[l]{Ours (Monolingual)}}
& \textbf{96.2} & 23.3 & \textbf{99.2} & 17.5 & 97.6 & 16.1 & 91.5 & 8.9 & 97.0 & 11.2 & \textbf{96.9} & 16.7 & 16.14 \\
& \textbf{\shortstack[l]{Ours (Parallel)}}
& 93.1 & 26.0 & 98.6 & 17.9 & 99.1 & 17.6 & \textbf{92.5} & \textbf{9.4} & \textbf{98.7} & 13.4 & 96.3 & 18.1 & \textbf{17.40} \\
\midrule
\multirow{2}{*}{Llama3.1-8b}
& \textbf{\shortstack[l]{Ours (Monolingual)}}
& \textbf{97.2} & \textbf{28.4} & \textbf{97.6} & \textbf{19.0} & \textbf{93.9} & \textbf{21.0} & \textbf{80.8} & \textbf{11.3} & \textbf{99.0} & \textbf{18.4} & \textbf{93.9} & \textbf{20.7} & \textbf{19.47} \\
& \textbf{\shortstack[l]{Ours (Parallel)}}
& 70.3 & 11.9 & 79.3 & 6.9 & 78.1 & 14.4 & 69.8 & 5.4 & 98.0 & 9.9 & 76.8 & 10.0 & 7.70 \\
\midrule
\multirow{2}{*}{Aya23-8b}
& \textbf{\shortstack[l]{Ours (Monolingual)}}
& 68.0 & \textbf{21.7} & \textbf{69.4} & \textbf{14.7} & \textbf{22.8} & 20.3 & 69.8 & 10.8 & 58.7 & 14.0 & 55.6 & 16.5 & 9.34 
\\
& \textbf{\shortstack[l]{Ours (Parallel)}}
& \textbf{88.0} & 20.1 & 58.8 & 14.4 & \textbf{22.8} & \textbf{25.6} & \textbf{76.0} & \textbf{12.8} & \textbf{80.0} & \textbf{16.1} & \textbf{61.7} & \textbf{17.3} & \textbf{10.69} \\
\bottomrule
\end{tabular}}
\caption{Result of multilingual generation control. The best performance for each metric is highlighted in bold. A*B stands for ACC*BLEU. 
Higher values indicate better performance.
Results are averaged over three runs with distinct random seeds used to sample sentences for identifying language-specific dimensions.}
\label{tab:mt_all}
\end{table*}

Additionally, we assess the consistency of the language-specific dimensions identified by the monolingual and parallel settings. As shown in Fig.~\ref{fig:overlap2}, when $K\!=\!400$ the average agreement across languages is 44.6\%. As $K$ is reduced to 50, the agreement increases to 77.6\%. This suggests that the two settings converge on the core language-specific dimensions.

\begin{CJK}{UTF8}{gbsn}
\begin{table*}[t]
\centering
\setlength{\arrayrulewidth}{0.6pt}        
\renewcommand{\arraystretch}{1.18}        
\begin{tabularx}{\textwidth}{|>{\raggedright\arraybackslash}p{1.5cm}|
                                 >{\raggedright\arraybackslash}p{4.2cm}|
                                 >{\raggedright\arraybackslash}X|}
\hline
\textbf{Input} & - & Translate an English sentence into a target language.\texttt{\textbackslash n}
English: \textbf{Why might someone prefer to shop at a small, locally-owned business instead of a large chain store, even if the prices are higher?}\texttt{\textbackslash n}
Target language: \\
\hline
\multirow{6}{*}{\textbf{Output}}
& \textbf{$\alpha\!=\!0.5$}
& 为什么一个人可能会选择在小型本地商店购物，而不是大型连锁商店？\textcolor{lightgray}{尽管价格更高}\textcolor{lightgray}{(even if the prices are higher)} \\
\cline{2-3}
& \textbf{$\alpha\!=\!0.7$}
& 为什么一个人可能会选择在小型本地商店购物，而不是大型连锁商店，尽管价格更高？ \\
\cline{2-3}
& \textbf{$\alpha\!=\!0.9$}
& 为什么一个人可能会选择在小型本地商店购物，而不是大型\textcolor{red}{链式}\textcolor{red}{(chain)}商店，尽管价格\textcolor{lightgray}{更}(high\textcolor{lightgray}{er})高？ \\
\cline{2-3}
\hline
\end{tabularx}
\caption{Language steering examples with different intervention strengths.
Examples are from Llama2–7B with an intervention at layer 27. They show how $\alpha$ affects generation.
Characters in \textcolor{lightgray}{gray} indicate content missing relative to the source sentence; characters in \textcolor{red}{red} indicate content that is mistranslated.}
\label{tab:examples}
\end{table*}
\end{CJK}

\begin{figure}[t]
    \centering
    \begin{subfigure}[t]{\linewidth}
        \centering
        \includegraphics[width=0.9\linewidth]{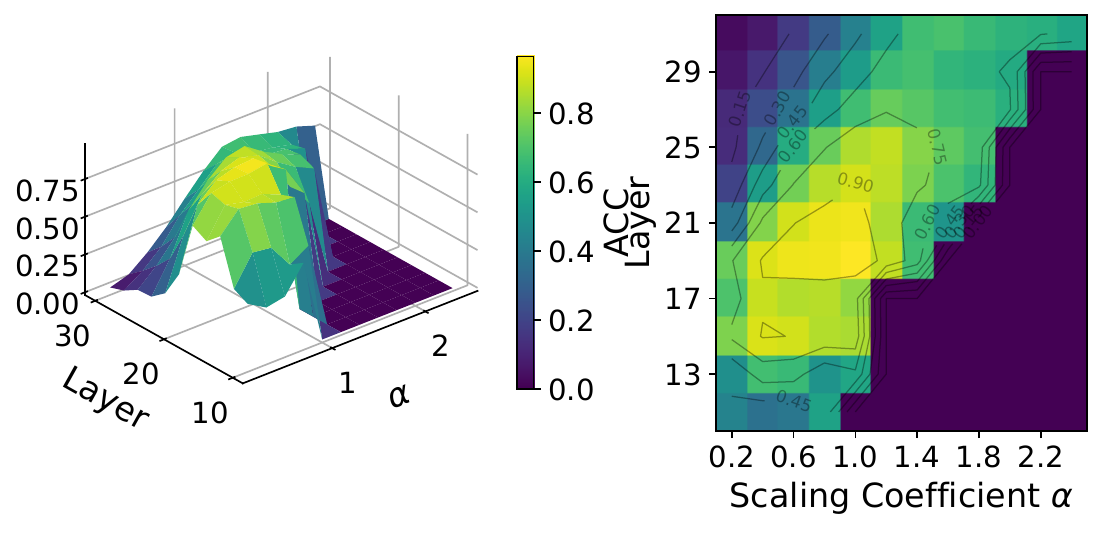}
        \caption{Overall ACC}
    \end{subfigure}

    
    \begin{subfigure}[t]{\linewidth}
        \centering
        \includegraphics[width=0.9\linewidth]{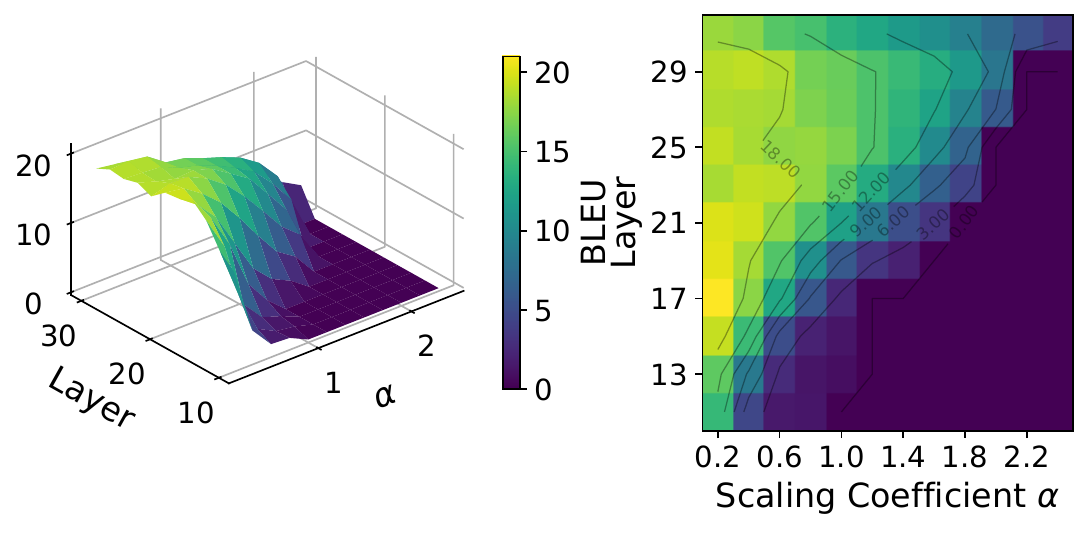}
        \caption{Overall BLEU}
    \end{subfigure}
    
    
    \begin{subfigure}[t]{\linewidth}
        \centering
        \includegraphics[width=0.9\linewidth]{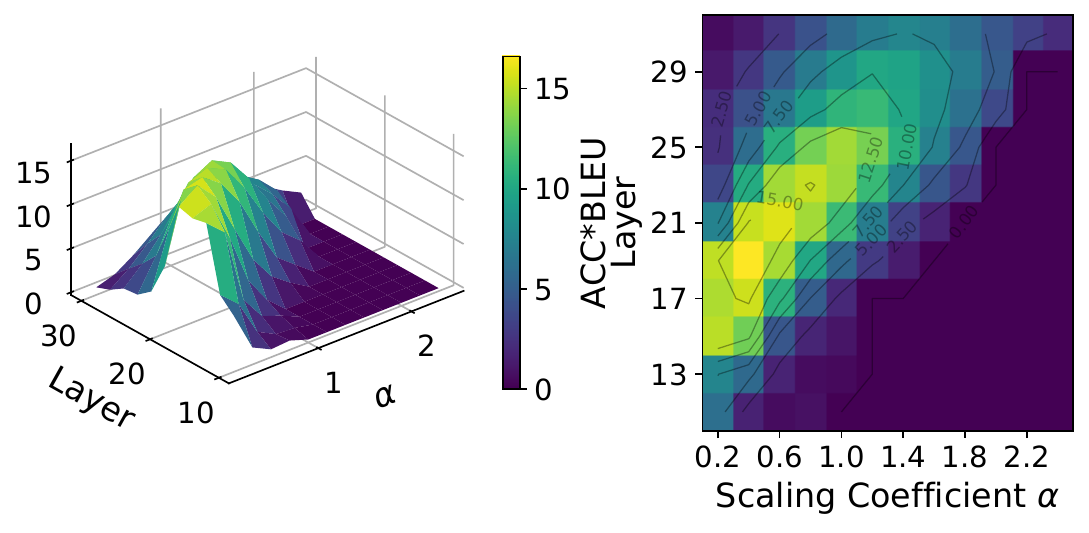}
        \caption{Overall ACC*BLEU}
    \end{subfigure}
    \caption{Grid search results of Llama2-7B in monolingual setting. Purple regions denote unevaluated configurations; their values are set to zero for visualization.} 
    \label{fig:grid_7b_mono}
\end{figure}

For the experiments in the later sections, as established in earlier results, we fix top-$K\!=\!400$ language-specific dimension identification for all models. In the monolingual setting, we compute differences between an intermediate layer and the final layer, using layer 20 for Llama2–7B, Llama3.1–8B, and Aya23–8B, and layer 22 for Llama2–13B. For the multilingual generation control task, we then apply single-layer inference-time interventions with scaling coefficient $\alpha$.
The intervention layer and $\alpha$ are selected via a grid search, hyper-parameter details are provided in Section~\ref{sec:grid} and Appendix~\ref{sec:details}. Table~\ref{tab:mt_all} reports the results obtained under the best settings for each model.

\subsection{Main Results of Controlling Multilingual Generation}
We validate our method on the multilingual generation control task as described in section~\ref{sec:task}. 
In this task, we compare our method with two neuron-based intervention baselines. Neuron-Kojima~\cite{neuron_kojima} ranks neurons by their ability to predict the target language from per sentence mean activations, then at inference replaces the selected neurons’ activations to steer the output. Neuron-Tang~\cite{neuron_tang} selects neurons with low cross language activation entropy after thresholding and controls generation by selectively turning these neurons on or off. We use the official code and settings for all experiments.

We show the results in Table~\ref{tab:mt_all}, evaluated on four LLMs across five language directions. Both variants of our approach can steer the models to produce the desired target language. Considering ACC*BLEU, our monolingual setting achieves an overall score of 16.63 on Llama2-7B, surpassing Neuron-Kojima's 7.76 by more than two times and 1.49 points above Neuron-Tang's 15.14. On Llama2-13B, it reaches 16.14, outperforming Neuron-Kojima by 11.43 points, and Neuron-Tang by 4.61 points. In the parallel setting, our method attains 16.57 on Llama2-7B, roughly comparable with the monolingual setting.
On Llama2-13B, the performance reaches 17.40, improving over the monolingual results by 1.26 points.

To assess generality, we evaluate on two latest models—Llama3.1–8B and Aya23–8B. On Llama3.1–8B in the monolingual setting, the overall ACC*BLEU reaches 18.70, the best among the settings we evaluated. The relatively lower performance of Llama3.1-8B under the parallel setting may be attributed to the use of hyperparameters tuned for Llama2-7B. On Aya23–8B in the parallel setting, it reaches 10.51. These results support our hypothesis about language-specific dimensions and indicate broad applicability across model families. 
We omit neuron-based baselines, as they are not compatible with these newer models. Besides BLEU, we additionally report BERTScore as an alternative semantic evaluation metric; the results are presented in Appendix~\ref{sec:bert}.

\subsection{Layerwise Analysis of Language-Specific Dimensions}\label{sec:grid}

To test our hypothesis that language-specific dimensions are consistent across layers, we apply an intervention to each intermediate layer while varying the scaling coefficient $\alpha$. We visualize the grid search results of Llama2-7B with monolingual setting in Fig.~\ref{fig:grid_7b_mono}. 
As shown in Fig.~\ref{fig:grid_7b_mono}, 
for most intermediate layers,
intervening on the same set of language-specific dimensions and choosing an appropriate $\alpha$ can yield a success rate exceeding 60\% of steering the output to the target language, supporting our hypothesis that language-specific dimensions are aligned across layers. 

\begin{CJK}{UTF8}{gbsn}
From the figure, we observe that achieving higher accuracy requires larger scaling coefficient $\alpha$ at later layers. This is because the magnitude of the representation increases with depth, so stronger interventions are needed to dominate the signal. In contrast, translation quality, measured by BLEU, tends to decrease as $\alpha$ increases. A plausible explanation is that strong interventions rigidly steer the model toward the target language, reducing its freedom to plan and revise the generation. The model then produces more literal, word-by-word translations with lower fluency. An illustrative example with Llama2–7B is shown in Table~\ref{tab:examples}. With the intervention applied at layer 27, at $\alpha\!=\!0.5$, the model omits the clause ``even if the prices are higher.'' At $\alpha\!=\!0.9$, the word ``chain'' is rendered as ``链式,'' whereas the idiomatic translation is ``连锁商店'' (chain store). At $\alpha\!=\!0.7$, we get a good translation. This illustrates that an appropriate $\alpha$ should be chosen to let the model translate into the right target language while preserving some freedom to produce more refined generation.\end{CJK} We also observe higher BLEU when the intervention is applied to earlier layers, indicating that early target-language signals allow later layers to devote more capacity to improving output quality.

\begin{table}[t]
\centering
\setlength{\tabcolsep}{2pt}
\renewcommand{\arraystretch}{1.5}
\resizebox{\columnwidth}{!}{
\begin{tabular}{ll *{2}{c} *{2}{c} *{2}{c} *{2}{c}}
\toprule
\textbf{Model} & \textbf{Method}
& \multicolumn{2}{c}{\textbf{De}}
& \multicolumn{2}{c}{\textbf{Zh}}
& \multicolumn{2}{c}{\textbf{Es}}
& \multicolumn{2}{c}{\textbf{Overall}} \\
\cmidrule(lr){3-4}\cmidrule(lr){5-6}\cmidrule(lr){7-8}\cmidrule(lr){9-10}
&
& \textbf{F1} & \textbf{EM}
& \textbf{F1} & \textbf{EM}
& \textbf{F1} & \textbf{EM}
& \textbf{F1} & \textbf{EM} \\
\midrule
\multirow{3}{*}{Llama2-7b}
& \textbf{Neuron Tang}
& \textbf{30.8} & \textbf{18.1}
& 6.8 & \textbf{6.8}
& \textbf{32.2} & \textbf{16.4}
& \textbf{23.3} & \textbf{13.8} \\
& \textbf{Ours (Monolingual)}
& 27.4 & 15.7
& \textbf{8.1} & 5.9
& 25.9 & 13.4
& 20.5 & 11.7 \\
& \textbf{Ours (Parallel)}
& 25.4 & 14.5
& 7.3 & 5.5
& 26.5 & 13.6
& 19.8 & 11.2 \\
\midrule
\multirow{3}{*}{Llama2-13b}
& \textbf{Neuron Tang}
& 30.2 & 18.6
& 10.4 & 8.9
& 27.4 & 14.8
& 22.7 & 14.1 \\
& \textbf{Ours (Monolingual)}
& 33.1 & \textbf{22.7}
& 10.8 & 9.3
& 34.7 & 17.0
& 26.2 & 16.4 \\
& \textbf{Ours (Parallel)}
& \textbf{33.9} & 20.2
& \textbf{11.7} & \textbf{10.0}
& \textbf{36.7} & \textbf{18.6}
& \textbf{27.4} & \textbf{16.3} \\
\bottomrule
\end{tabular}}
\caption{Results on MLQA. For reference, the English QA baseline without any intervention achieves F1/EM of 51.1/30.7 on Llama2-7B and 68.2/48.4 on Llama2-13B. All reported results use English prompts with interventions and are evaluated against gold answers in the target language.}
\label{tab:MLQA_all}
\end{table}

\subsection{Results of MLQA}\label{sec:MLQA}

We further evaluate the proposed intervention methods on MLQA, which constitutes a more challenging setting that requires the model to correctly follow instructions, comprehend the given context, and accurately locate the gold answer span.

The results are summarized in Table~\ref{tab:MLQA_all}.
Overall, our method achieves performance comparable to the neuron-based approach.
While the neuron-based method performs slightly better on the 7B model, our approach outperforms it on the 13B model.
This trend may be related to the increased stability of internal representations in larger models.
However, all intervention-based methods perform substantially worse than the base model without intervention.

Beyond the effect of the intervention itself, this degradation also arises from the extractive nature of MLQA, where the model must generate target-language answers that do not appear in the English context.
As a result, the task effectively shifts from extractive QA to cross-lingual answer generation, which inherently increases difficulty.

Nevertheless, both neuron-level and hidden-state-level intervention methods still demonstrate a non-trivial ability to induce language switching.
This suggests that for more complex tasks such as QA, the model’s internal behavior becomes considerably more intricate: representation-level manipulations can influence the output language, albeit with a trade-off in overall task performance.

\section{Conclusion}

In this work, we investigate how LLMs encode language identity by tracing the trajectory of representations across layers. We show that the transition from a English-aligned representation space to target-language token spaces is governed by a small and sparse set of language-specific dimensions that are consistent across layers. We introduce a simple, training-free method to identify and manipulate these dimensions, using as few as 50 sentences.
Our inference-time interventions demonstrate that manipulating these dimensions enables controlled language switching while preserving semantic content, thereby revealing their interpretability and practical utility. Experimental results on multilingual generation control task confirm the effectiveness of our approach: it achieves strong performance in both monolingual and parallel settings, surpassing Neuron-Kojima by up to 12.69 points and Neuron-Tang by up to 5.87 points. Moreover, the parallel setting is generally stronger than the monolingual setting.
These findings establish language-specific dimensions as a consistent and controllable mechanism underlying cross-lingual transfer in LLMs, opening new directions for both interpretability and efficient multilingual control.


\section{Limitations}
Most ablations, such as anchor layer, top-K, are conducted on Llama2–7B and then transferred to Llama3.1–8B and Aya23–8B. Although these models have similar depths, the optimal hyperparameters may differ by model. As shown in Table~\ref{tab:mt_all}, in the parallel setting, the performance of Llama3.1–8B is markedly lower than in the monolingual setting, suggesting that the transferred hyperparameters are suboptimal for this model/setting. A systematic per-model search is left to future work.

Our empirical evaluation primarily focuses on multilingual generation control under a specific prompt design, with a limited study on question answering.
While the results demonstrate the effectiveness of our method in controlled settings, its impact on a broader range of downstream tasks, such as mathematics and summarization, remains to be systematically examined.
Future work should extend the evaluation scope to diverse tasks, incorporate task-specific metrics and human judgments, and assess robustness across different prompt formulations.

\section*{Acknowledgments}
This work was supported by the ``R\&D Hub Aimed at Ensuring Transparency and Reliability of Generative AI Models'' project of the Ministry of Education, Culture, Sports, Science and Technology, and by JSPS KAKENHI Grant Number JP23K28144, and by the Cross-ministerial Strategic Innovation Promotion Program (SIP) on ``Integrated Health Care System'' Grant No. JPJ012425.

\nocite{*}
\bibliography{custom}


\appendix

\section{Details of Intervention Setting}\label{sec:details}
\begin{figure}[h]
  \centering
  \includegraphics[width=0.8\linewidth]{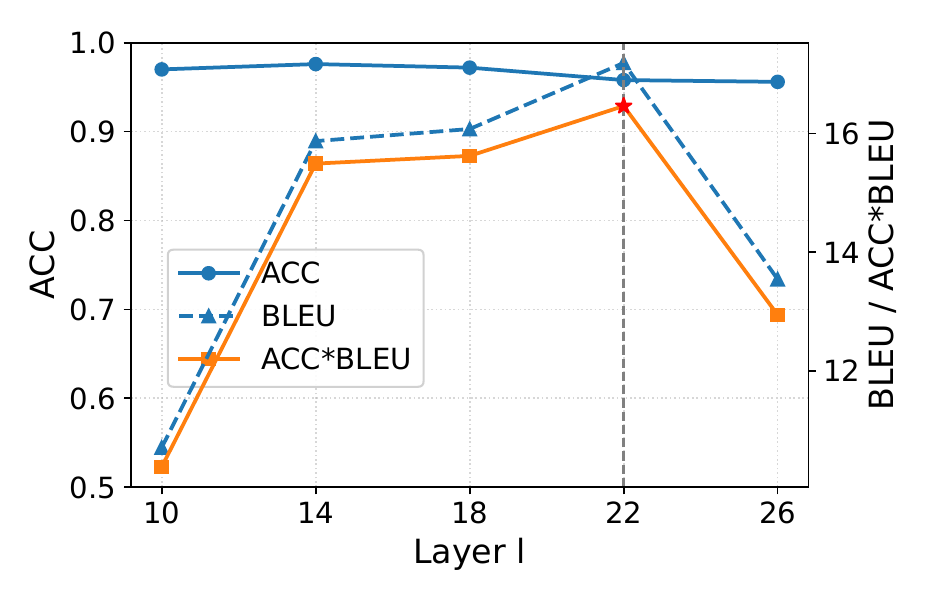}
  \caption{Choosing an anchor layer for monolingual setting. Ablation on Llama2–13B.}
  \label{fig:inter_llama13b}
\end{figure}

The intervention configurations used to obtain the results in Table~\ref{tab:mt_all} are: Llama2–7B—layer 19, $\alpha\!=\!0.4$ for monolingual and parallel; Llama2–13B—layer 19, $\alpha\!=\!0.4$ for monolingual and $\alpha\!=\!0.3$ for parallel; Llama3.1–8B—layer 21, $\alpha\!=\!0.4$ for monolingual and layer 27, $\alpha\!=\!0.8$ for parallel; Aya23–8B—layer 29, $\alpha\!=\!1.2$ for monolingual and layer 27, $\alpha\!=\!0.8$ for parallel.
The intervention layer and $\alpha$ are selected via a grid search

\subsection{Choosing an anchor layer for Llama2-13B}
Ablation on Llama2–13B shows that using layer 22 as the intermediate reference and contrasting it with the final layer yields the most effective set of language-specific dimensions.

\subsection{Extra Experiments}
We present grid search results over the intervention layer and the scaling coefficient $\alpha$ for Llama2–7B in parallel setting, and Llama2–13B in monolingual and parallel settings.

Based on the grid search, the optimal intervention settings are as follows: for Llama2–7B under the parallel setting, layer 19 with $\alpha\!=\!0.4$; for Llama2–13B under the monolingual setting, layer 19 with $\alpha\!=\!0.4$; for Llama2–13B under the parallel setting, layer 19 with $\alpha\!=\!0.3$.

\begin{figure}[t]
    \centering
    \begin{subfigure}[t]{\linewidth}
        \centering
        \includegraphics[width=0.9\linewidth]{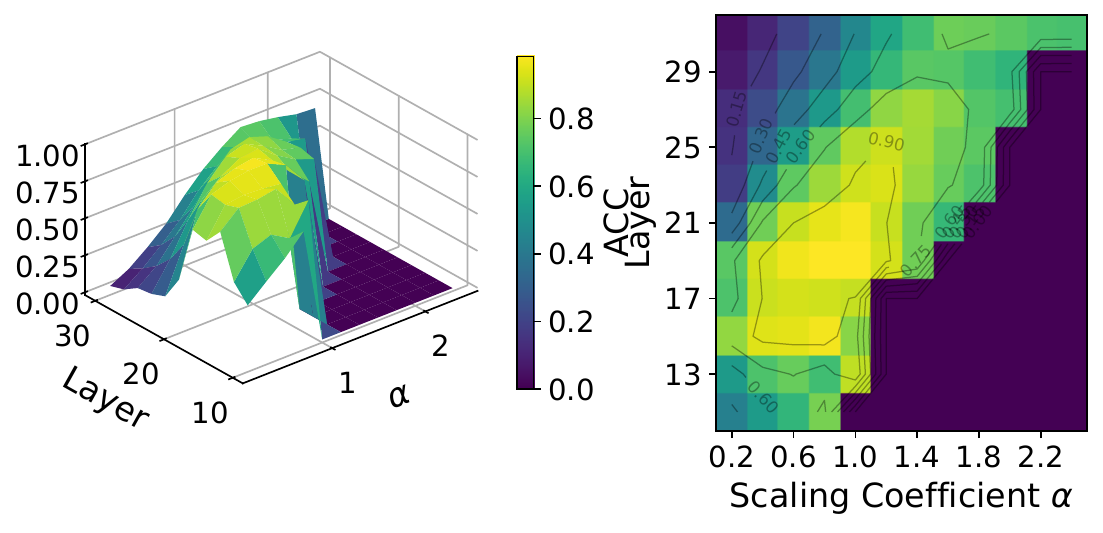}
        \caption{Overall ACC}
    \end{subfigure}

    
    \begin{subfigure}[t]{\linewidth}
        \centering
        \includegraphics[width=0.9\linewidth]{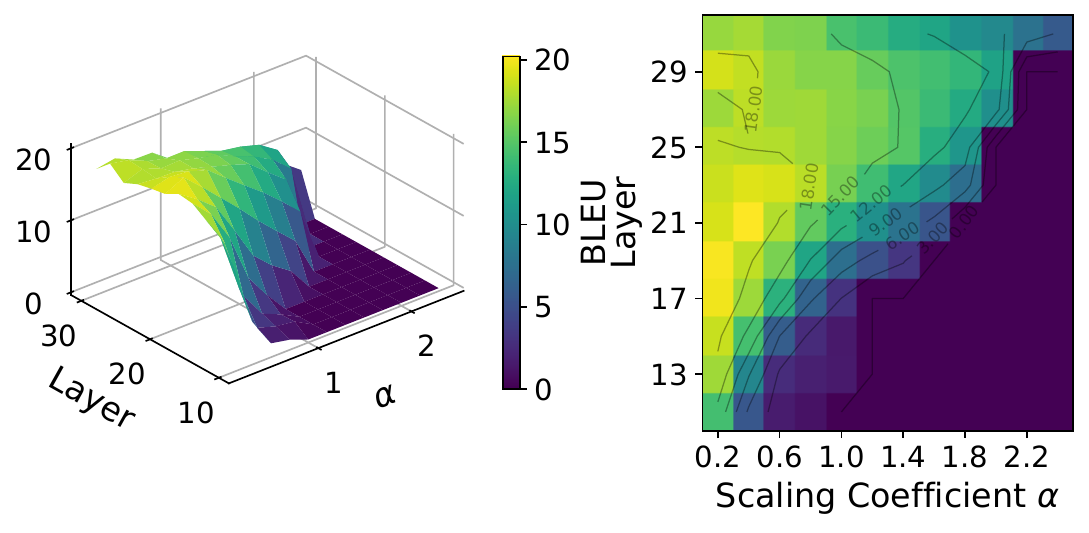}
        \caption{Overall BLEU}
    \end{subfigure}
    
    
    \begin{subfigure}[t]{\linewidth}
        \centering
        \includegraphics[width=0.9\linewidth]{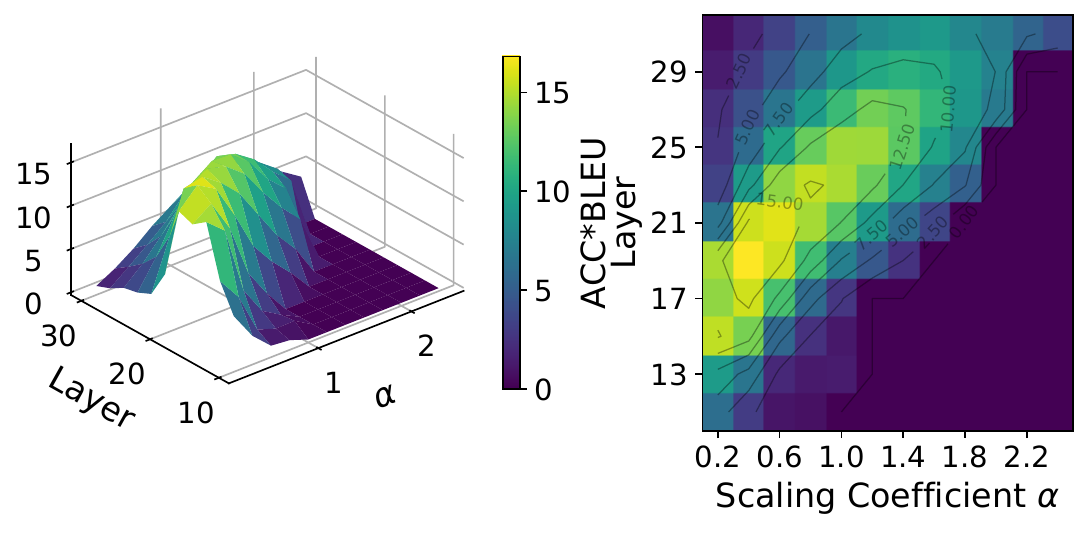}
        \caption{Overall ACC*BLEU}
    \end{subfigure}
    \caption{Grid search results of Llama2-7B in parallel setting. The results show that the most effective intervention should be applied to layer 19, with $\alpha\!=\!0.4$.} 
    \label{fig:grid}
\end{figure}

\begin{figure}[!h]
  \centering
  \includegraphics[width=0.8\linewidth]{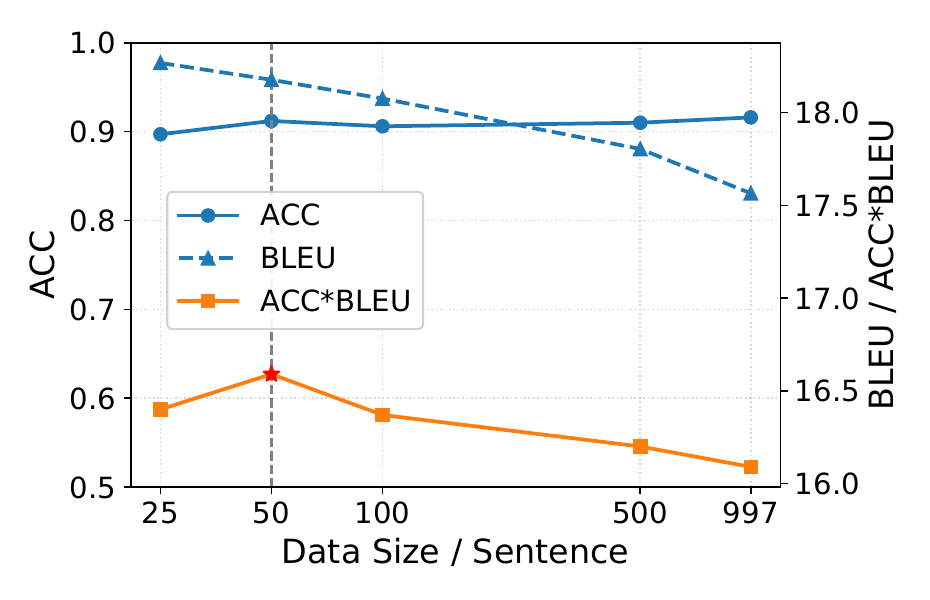}
  \caption{Identify Language-specific Dimensions. Ablation study on Llama2-7B in the monolingual setting.}
  \label{fig:data}
\end{figure}

\subsection{Are 50 Sentences Sufficient for Identification}
When identifying language-specific dimensions, we randomly sample 50 parallel sentence pairs from the FLORES-200. While using a small corpus reduces data and compute requirements, it can also introduce greater stochasticity. 
To assess whether larger data samples could reduce stochasticity and improve the accuracy of language-specific dimension identification, we conduct experiments with increased corpus sizes. The results are shown in Fig.~\ref{fig:data}.
The experiments are conducted on Llama2-7B, with an intervention applied at layer 20  with $\alpha\!=\!0.4$.
As corpus size increases, ACC*BLEU performance does not improve. This indicates that our method achieves strong performance with minimal data, obviating the need for large corpora.

\begin{table*}[!ht]
\centering
\setlength{\tabcolsep}{2pt}
\renewcommand{\arraystretch}{1.5}
\resizebox{\textwidth}{!}{
\begin{tabular}{ll *{2}{c} *{2}{c} *{2}{c} *{2}{c} *{2}{c} *{3}{c}}
\toprule
\textbf{Model} & \textbf{Method}
& \multicolumn{2}{c}{\textbf{Fr}}
& \multicolumn{2}{c}{\textbf{De}}
& \multicolumn{2}{c}{\textbf{Zh}}
& \multicolumn{2}{c}{\textbf{Ja}}
& \multicolumn{2}{c}{\textbf{Es}}
& \multicolumn{3}{c}{\textbf{Overall}} \\
\cmidrule(lr){3-4}\cmidrule(lr){5-6}\cmidrule(lr){7-8}\cmidrule(lr){9-10}\cmidrule(lr){11-12}\cmidrule(lr){13-15}
&
& \textbf{ACC} & \textbf{BERT}
& \textbf{ACC} & \textbf{BERT}
& \textbf{ACC} & \textbf{BERT}
& \textbf{ACC} & \textbf{BERT}
& \textbf{ACC} & \textbf{BERT}
& \textbf{ACC} & \textbf{BERT} & \textbf{A*B} \\
\midrule
\multirow{3}{*}{Llama2-7b}
& \textbf{\shortstack[l]{Neuron Tang}}
& 83.7 & 0.945 & 72.7 & \textbf{0.941} & 86.7 & \textbf{0.918} & 20.5 & \textbf{0.922} & 98.0 & \textbf{0.945} & 72.3 & \textbf{0.935} & 67.58 \\
& \textbf{\shortstack[l]{Ours (Monolingual)}}
& 98.3 & \textbf{0.952} & 97.3 & 0.939 & 84.7 & 0.912 & \textbf{76.5} & 0.904 & 98.7 & 0.940 & 91.1 & 0.929 & 84.63 \\
& \textbf{\shortstack[l]{Ours (Parallel)}}
& \textbf{99.1} & 0.940 & \textbf{98.8} & 0.938 & \textbf{88.3} & 0.916 & \textbf{76.5} & 0.905 & \textbf{99.3} & 0.940 & \textbf{92.6} & 0.930 & \textbf{86.02} \\
\midrule
\multirow{3}{*}{Llama2-13b}
& \textbf{\shortstack[l]{Neuron Tang}}
& 83.7 & \textbf{0.948} & 14.7 & \textbf{0.942} & \textbf{99.3} & \textbf{0.925} & 64.5 & 0.896 & 42.0 & \textbf{0.949} & 63.7 & \textbf{0.936} & 59.62 \\
& \textbf{\shortstack[l]{Ours (Monolingual)}}
& \textbf{96.2} & 0.936 & \textbf{99.2} & 0.935 & 97.6 & 0.909 & 91.5 & \textbf{0.907} & 97.0 & 0.907 & \textbf{96.9} & 0.921 & \textbf{89.27} \\
& \textbf{\shortstack[l]{Ours (Parallel)}}
& 93.1 & 0.940 & 98.6 & 0.934 & 99.1 & 0.914 & \textbf{92.5} & 0.903 & \textbf{98.7} & 0.916 & 96.3 & 0.924 & 88.96 \\
\bottomrule
\end{tabular}}
\caption{Results of multilingual generation control with semantic preservation evaluated by XLM-R.}
\label{tab:mt_bert}
\end{table*}

\subsection{Results of BERTScore}\label{sec:bert}
In addition to BLEU, we also report semantic similarity using BERTScore with FacebookAI/xlm-roberta-large. As shown in Table~\ref{tab:mt_bert}, all methods exhibit comparable semantic preservation, with substantially smaller performance gaps than those observed with BLEU. Since BERTScore is evaluated only on samples where the generated output is in the correct target language, the improvement in the ACC*BERTScore metric mainly reflects the higher language accuracy achieved by our method, resulting in a clear advantage over the neuron-based baseline.

\begin{figure}[ht]
    \centering
    \begin{subfigure}[t]{\linewidth}
        \centering
        \includegraphics[width=0.9\linewidth]{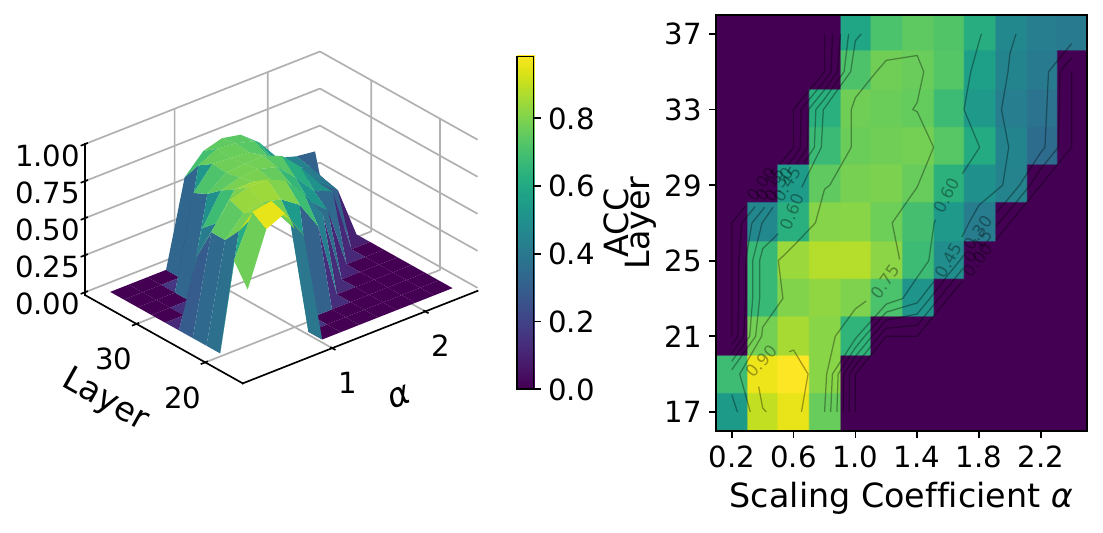}
        \caption{Overall ACC}
    \end{subfigure}

    
    \begin{subfigure}[t]{\linewidth}
        \centering
        \includegraphics[width=0.9\linewidth]{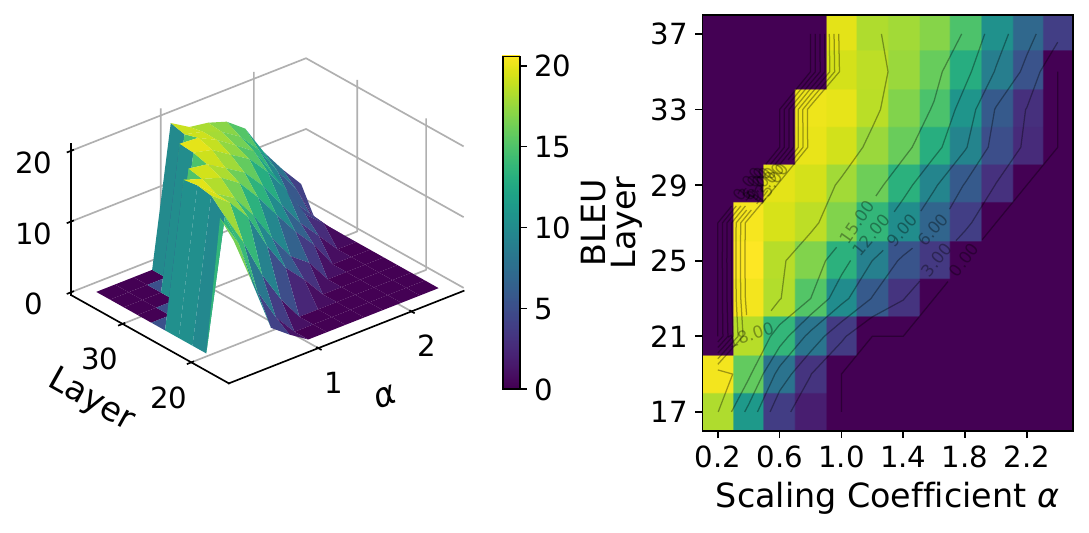}
        \caption{Overall BLEU}
    \end{subfigure}
    
    
    \begin{subfigure}[t]{\linewidth}
        \centering
        \includegraphics[width=0.9\linewidth]{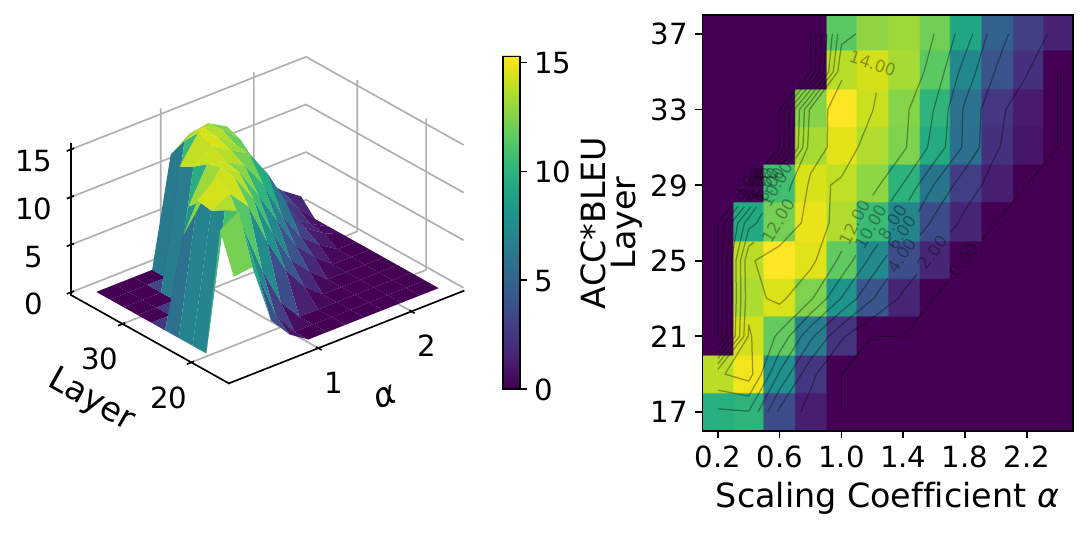}
        \caption{Overall ACC*BLEU}
    \end{subfigure}
    \caption{Grid search results of Llama2-13B in monolingual setting. The results show that the most effective intervention should be applied to layer 19, with $\alpha\!=\!0.4$.} 
    \label{fig:grid_llama13b_mono}
\end{figure}

\begin{figure}[h]
    \centering
    \begin{subfigure}[t]{\linewidth}
        \centering
        \includegraphics[width=0.9\linewidth]{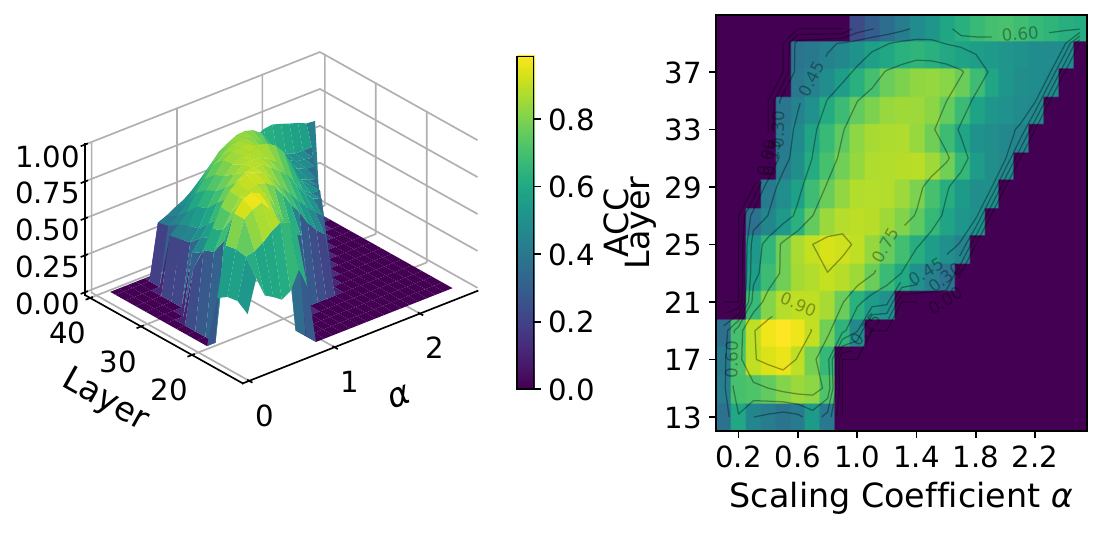}
        \caption{Overall ACC}
    \end{subfigure}

    
    \begin{subfigure}[t]{\linewidth}
        \centering
        \includegraphics[width=0.9\linewidth]{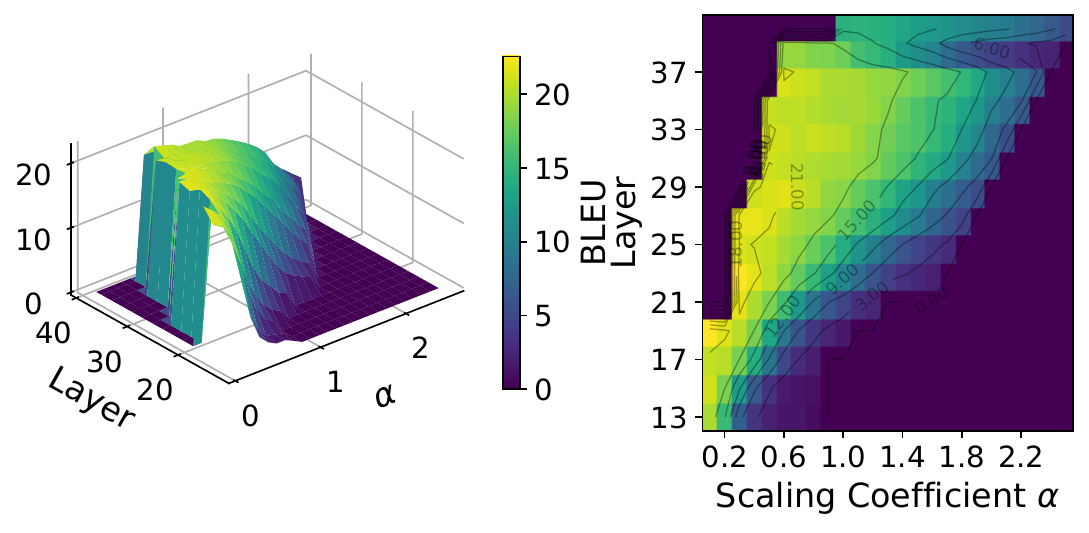}
        \caption{Overall BLEU}
    \end{subfigure}
    
    
    \begin{subfigure}[t]{\linewidth}
        \centering
        \includegraphics[width=0.9\linewidth]{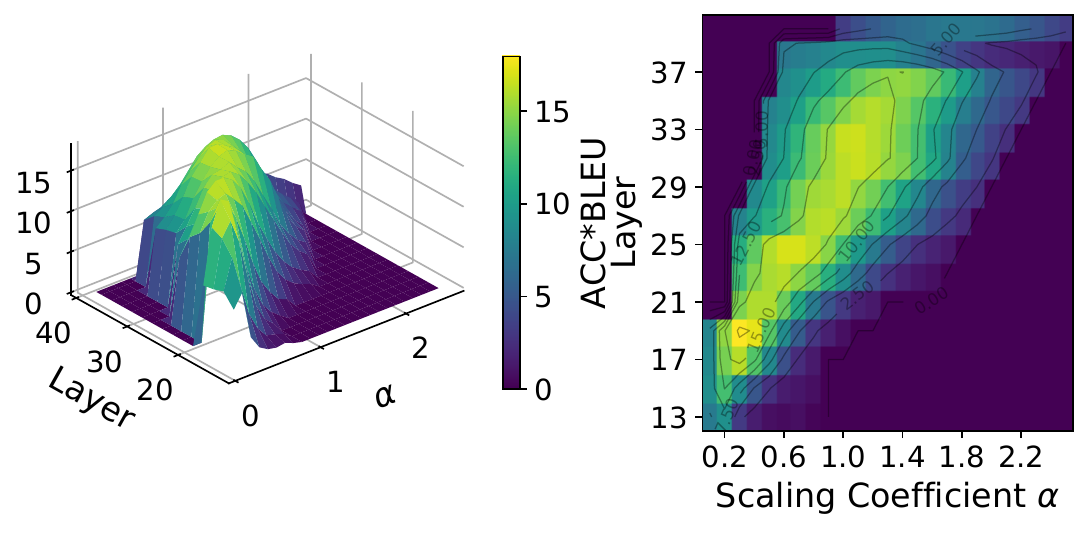}
        \caption{Overall ACC*BLEU}
    \end{subfigure}
    \caption{Grid search results of Llama2-13B in parallel setting. The results show that the most effective intervention should be applied to layer 19, with $\alpha\!=\!0.3$.} 
    \label{fig:grid_llama13b_para}
\end{figure}

\end{document}